\documentclass{article}

\usepackage{arxiv}

\usepackage[utf8]{inputenc} 
\usepackage[T1]{fontenc}    
\usepackage{hyperref}       
\usepackage{url}            
\usepackage{booktabs}       
\usepackage{amsfonts}       
\usepackage{nicefrac}       
\usepackage{microtype}      
\usepackage{lipsum}         
\usepackage{graphicx}
\usepackage{natbib}
\usepackage{doi}

\usepackage{amsmath}
\usepackage{amssymb}
\usepackage{stmaryrd}
\usepackage{xcolor}
\usepackage{enumerate}
\usepackage{mathtools}
\usepackage{subcaption}
\usepackage{booktabs}
\usepackage{multirow}
\usepackage{tabularx}

\newtheorem{definition}{Definition}

\renewcommand{\vec}[1]{\boldsymbol{#1}}
\newcommand{\given}{\, | \,}

\newcommand{\vx}{{\vec{x}}}
\newcommand{\vtheta}{{\vec{\theta}}}
\newcommand{\sumK}{\sum_{k=1}^K}
\newcommand{\bE}{\mathbb{E}}

\newcommand{\fromto}{\longrightarrow}

\newcommand*{\defeq}{\mathrel{\vcenter{\baselineskip0.5ex \lineskiplimit0pt
			\hbox{\footnotesize.}\hbox{\footnotesize.}}}%
	=}

\newcommand{\cY}{\mathcal{Y}}
\newcommand{\cH}{\mathcal{H}}
\newcommand{\cD}{\mathcal{D}}
\newcommand{\eu}{\operatorname{EU}}
\newcommand{\au}{\operatorname{AU}}

\newcommand{\argmax}{\operatorname*{argmax}}
\newcommand{\on}[1]{\operatorname{#1}}

\newcommand{\newinf}{\mathop{\mathrm{inf}\vphantom{\mathrm{sup}}}}

\definecolor{darkgreen}{rgb}{0.0, 0.6, 0.0} 

\definecolor{darkred}{rgb}{0.9, 0.0, 0.0} 

\title{Quantifying Aleatoric and Epistemic Uncertainty \\ with Proper Scoring Rules}

\author{Paul~Hofman \\
	Institute of Informatics, LMU Munich\\
	Munich Center for Machine Learning (MCML)\\
	\texttt{paul.hofman@ifi.lmu.de} \\
  	\And
	Yusuf~Sale \\
    Institute of Informatics, LMU Munich\\
	Munich Center for Machine Learning (MCML)\\
	\texttt{yusuf.sale@ifi.lmu.de} 
	\And
    Eyke~Hüllermeier\\
    Institute of Informatics, LMU Munich\\
	Munich Center for Machine Learning (MCML)\\
	\texttt{eyke@lmu.de} \\
}

\renewcommand{\shorttitle}{Quantifying Aleatoric and Epistemic Uncertainty with Proper Scoring Rules}

\hypersetup{
pdftitle={Quantifying Aleatoric and Epistemic Uncertainty with Proper Scoring Rules},
pdfsubject={stat.ML},
pdfauthor={Paul Hofman, Yusuf Sale, Eyke Hüllermeier},
pdfkeywords={Uncertainty quantification, credal sets, second-order distributions, proper scoring rules},
}

\begin{document}
\maketitle

\begin{abstract}
    Uncertainty representation and quantification are paramount in machine learning and constitute an important prerequisite for safety-critical applications. In this paper, we propose novel measures for the quantification of aleatoric and epistemic uncertainty based on proper scoring rules, which are loss functions with the meaningful property that they incentivize the learner to predict ground-truth (conditional) probabilities. We assume two common representations of (epistemic) uncertainty, namely, in terms of a credal set, i.e. a set of probability distributions, or a second-order distribution, i.e., a distribution over probability distributions. Our framework establishes a natural bridge between these representations. We provide a formal justification of our approach and introduce new measures of epistemic and aleatoric uncertainty as concrete instantiations.

\end{abstract}


\section{Introduction}
\label{sec:intro}
Understanding and handling uncertainty is a fundamental challenge in machine learning (ML) and artificial intelligence (AI) research. Due to the intrinsic complexities and variability of real-world data, coupled with the probabilistic nature of many ML algorithms, the latter are often subject to various forms of uncertainty. Unless properly addressed, this uncertainty can pose substantial limitations to the reliability of ML systems,
which is especially problematic in applications with stringent safety considerations, such as in medicine and the healthcare sector \citep{lambrou2010reliable, senge_2014_ReliableClassificationLearning, yang2009using}.
In these and other domains, the deployment of ML comes with the urgent need for adequate methods of representing and quantifying uncertainty.

In the broader scope of the literature, a distinction between \textit{aleatoric} and \textit{epistemic} uncertainty  is usually made \citep{hora1996aleatory}. Aleatoric uncertainty originates from the inherent stochastic nature of the data-generating process, while epistemic uncertainty is due to the learner's incomplete knowledge of this process. The latter can therefore be reduced by the acquisition of additional information, such as additional observations. Conversely, aleatoric uncertainty, being a characteristic of the data-generating process itself, is non-reducible \citep{hullermeier2021aleatoric}. Representing and quantifying these types of uncertainties have become pivotal in recent ML research, including Bayesian deep learning \citep{depeweg2018decomposition, kendall2017uncertainties}, adversarial example detection \citep{smith2018understanding}, and data augmentation in Bayesian classification \citep{kapoor2022uncertainty}.

The Bayesian approach, in which the learner's epistemic uncertainty is represented in terms of a probability distribution on the underlying model class, which in turn induces a second-order distribution on outcomes, prevails the ML literature so far. Uncertainty quantification for representations of that kind is commonly based on information-theoretic notions such as (Shannon) entropy and mutual information. Yet, this approach is not without criticism \citep{wimmer2023quantifying}. In this paper, we address this criticism by proposing alternative, more general measures of uncertainty based on proper scoring rules. In addition, we extend this approach to the notion of \emph{credal sets}, i.e., sets of probability distributions. Here, we assume a learner that produces (second-order) predictions in the form of sets of probability distributions on outcomes. For this representation, we also introduce new measures of epistemic and aleatoric uncertainty, thereby providing a flexible, unified framework to quantify uncertainty in different settings.

Our point of departure is a taxonomy of different types of uncertainty-aware learning algorithms introduced in the next section. In Section \ref{sec:uq}, we address the problem of uncertainty quantification and recall basic uncertainty measures for the probabilistic, Bayesian and Levi agent. We propose a novel construction of uncertainty measures for Bayesian and Levi agents based on (strictly) proper scoring rules in Section \ref{sec:psr} prior to concluding the paper in Section \ref{sec:conc}.

\section{Uncertainty  Representation}
\label{sec:uncertainty}

Throughout this paper, we consider a formal setting of supervised learning for categorical prediction (classification) as follows: 
Let $\mathcal{X}$ be a (measurable) instance space and $\mathcal{Y}$ a label space, where $\mathcal{Y} = \{ 1, \ldots, K \}$ for some $K \in \mathbb{N}$. Further, $\mathcal{D} = \{ ({x}_i, y_i )  \}_{i = 1}^n \in (\mathcal{X} \times \mathcal{Y})^n$ is a set of training data. The pairs $(x_i, y_i)$ are realizations of random variables $(X_i, Y_i)$, which are assumed to be independent and identically distributed (i.i.d.) according to some probability measure $P$ on $\mathcal{X} \times \mathcal{Y}$. We also assume a hypothesis space $\mathcal{H}$ to be given, where each hypothesis $h \in \mathcal{H}$ is a probabilistic predictor $\mathcal{X} \fromto \mathbb{P}(\mathcal{Y})$ that map instances $\vec{x}$ to a probability measure on $\mathcal{Y}$. 

Given $\mathcal{D}$ and $\mathcal{H}$, the learner induces a hypothesis $h$, the predictions $h(\vx) = \vtheta_\vx$ of which are considered as estimations of the ground-truth conditional distribution on $\mathcal{Y}$ given $X= \vx$, denoted $\vtheta^*_\vx$. For simplicity, we will often omit the (query) instance $\vx$ as a subscript. 
As $\mathbb{P}(\mathcal{Y})$, the set of all probability measures on $\mathcal{Y}$, can be identified with the $(K-1)$-simplex $\Delta_K$, both the estimate and the ground truth can be considered as elements of this simplex, i.e., as vectors $\vtheta = (\theta_1, \ldots , \theta_K)^\top$ and $\vtheta^* = (\theta_1^*, \ldots , \theta_K^*)^\top$, respectively, where $\theta_k^*$ is the true probability $P(Y = k \, | \, X = \vx)$ of the $k^{th}$ class (given $\vec{x}$) and $\theta_k$ the estimate of this probability.

\begin{figure}[t!]
    \centering
    \includegraphics[width=0.9\linewidth]{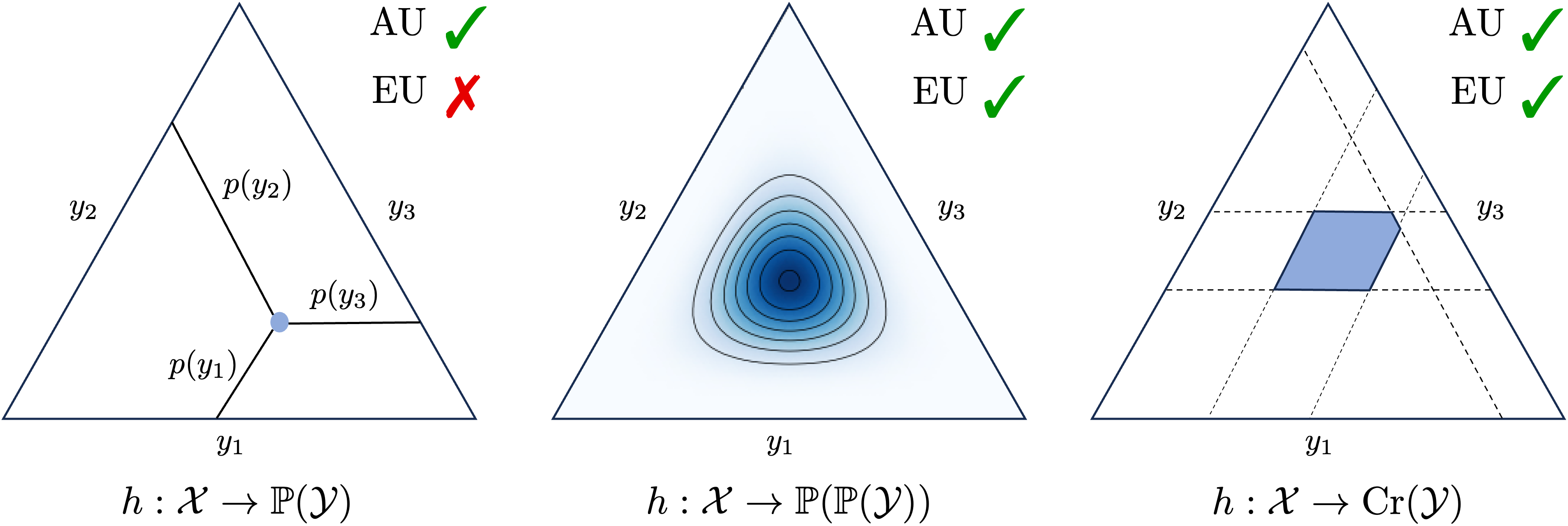}
    \caption{Uncertainty awareness in multi-class classification, illustrated on the probability simplex for $\mathcal{Y} = \{y_1, y_2, y_3\}$. From \textit{left} to \textit{right}: Probabilistic agent ($\on{AU}$, but \textit{\textcolor{darkred}{no}} $\on{EU}$ awareness), Bayesian agent ($\on{AU}$ \textit{\textcolor{darkgreen}{and}} $\on{EU}$ awareness), and Levi agent ($\on{AU}$ \textit{\textcolor{darkgreen}{and}} $\on{EU}$ awareness).}
    \label{fig:representations}
\end{figure}

Given a query $\vec{x} \in \mathcal{X}$ for which a prediction is sought, different learning methods proceed on the basis of different types of information. Depending on how the predictive uncertainty, i.e., the uncertainty about a predicted probability distribution $\vtheta \in \Delta_K$, is represented, we propose to distinguish four types of possible learners. Referring to the literature on decision under uncertainty, we call these learners deterministic agents, probabilistic agents, Bayesian agents, and Levi agents, respectively. An overview (of the last three) can be found in Figure \ref{fig:representations}.

\subsection{Deterministic Agents}
Very simple machine learning methods produce predictors in the form of mappings $\mathcal{X} \fromto \mathcal{Y}$, which assign to each query instance a predicted class label in a deterministic way. Obviously, a learner of that kind does not exhibit any sort of uncertainty-awareness, and does neither reflect aleatoric nor epistemic uncertainty.

\subsection{Probabilistic Agents}
A more common practice in contemporary machine learning is to consider probabilistic learners, which induce a hypothesis $h \in \cH$ and use this hypothesis to make predictions. Given $\vx \in \mathcal{X}$ as input, such a learner will predict a single probability distribution $h(\vx) = \vtheta$,
which is considered as an estimation of the ground-truth conditional probability $\vtheta^*$. We call a learner of that kind a \emph{probabilistic agent}. As it fully commits to a single $h \in \cH$, such an agent's uncertainty about the outcome $y$ is purely aleatoric. At the level of the hypothesis space, the agent pretends full certainty, and hence the absence of any epistemic uncertainty.

\subsection{Bayesian Agents}
Adhering to the principle of (strict) Bayesianism as advocated by statisticians such as De Finetti \citep{defi_fi80}, 
a \emph{Bayesian agent} will represent its belief in terms of a probability distribution on $\cH$. Thus, instead of committing to a single hypothesis, the agent will assign a probability (density) $q(h)$ to each candidate $h \in \cH$. Moreover, belief revision in the light of observed data $\cD$ is accomplished by replacing this distribution with the posterior $q(h \given \cD)$.

Since every $h \in \cH$ gives rise to a different probabilistic prediction $h(\vx)$, a Bayesian agent's belief about the outcome $y \in \mathcal{Y}$ is represented by a second-order distribution, i.e., a probability distribution of probability distributions. Formally, this distribution is the image of $q$ under the mapping $\mathcal{H} \fromto \Delta_K, h \mapsto h(\vx)$:
\begin{equation}\label{eq:qp}
p(\vtheta) = \int_{\mathcal{H}} \llbracket h(\vec{x}) = \vtheta \rrbracket \, d \, q(h \given \cD)  \, .
\end{equation}
Thus, $p(\vtheta)$ is the probability (density) of the probabilistic prediction $\vtheta$. 

In addition to the second-order distribution $p$, called posterior predictive distribution, a Bayesian agent typically also induces a representative (first-order) distribution via Bayesian model averaging:
\begin{equation*}\label{eq:bma}
\bar{\vtheta} \defeq \operatorname{bma}(p) = \int_{\cH} h(\vec{x})  \, d \, q(h \given \cD)  \, .
\end{equation*}

\subsection{Levi Agents}
\label{sub:levi}
Instead of representing model uncertainty in terms of a distribution on $\mathcal{H}$, this uncertainty could also be characterised in terms of an arguably even simpler model, namely, a subset $Q \subseteq \cH$ of hypotheses. According to this model, each $h \in Q$ is deemed a possible candidate predictor, whereas all $h \not\in Q$ are excluded as being implausible. Under the mapping $h \mapsto h(\vx)$, the set $Q$ directly translates into a set $C \subseteq \Delta_K$ of possible class distributions:
\begin{equation*}
C = \{ \vtheta = h(\vx) \given h \in Q \} \, .
\end{equation*}
In the literature, such a set of probability distributions is also referred to as a \emph{credal set} \citep{levi_te, wall_sr}. The reasonableness of taking decisions on the basis of sets of probability distributions (and thus deviating from strict Bayesianism) has been advocated by decision theorists like Isaac Levi \citep{levi_oi74,levi_te}. Correspondingly, we call a learner of this kind a \emph{Levi agent}. 

In the realm of machine learning, an approach of this kind is somewhat in line with the model of version space learning \citep{mitchell1977version}, i.e., the subset $Q$ can be seen as a kind of version space. From an uncertainty representation point of view, a set $Q \subseteq \cH$ appears to provide weaker information compared to a distribution $q \in \mathbb{P}(\cH)$. While this is true in a sense, a set-based representation does also have advantages. In particular, many have argued that probability distributions are less suitable for representing \emph{ignorance} in the sense of a lack of knowledge \citep{dubo_rp96}. For example, if the uniform distribution is taken as a model of \emph{complete ignorance}, as commonly done in probability theory, it is no longer possible to distinguish between a complete lack of knowledge and precise knowledge about the equal probability of all outcomes. Apart from that, one has to admit that the specification of a meaningful (prior) distribution is a difficult task in a machine learning setting, where $\cH$ is a very complex space.

\newpage
Another problem of a (second-order) probabilistic model is caused by the measure-theoretic grounding and additive nature of probability, which implies that the uniform distribution is not invariant under nonlinear transformation. As a consequence, even when starting with a uniform distribution on $\cH$, suggesting complete lack of knowledge about the right predictor, the image (\ref{eq:qp}) will normally not be uniform on $\Delta_K$. In other words, even if the learner is supposedly ignorant about the right predictor, it will pretend a certain degree of informedness about the prediction in a point $\vec{x} \in \mathcal{X}$. Even worse, different predictive distributions (and degrees of uncertainty) will be obtained for different instances $\vec{x} \in \mathcal{X}$.

\section{Uncertainty Quantification}
\label{sec:uq}
According to our discussion so far, different types of learners represent their information or ``belief'' about the outcome $y$ for an instance $\vec{x}$ in different ways, e.g., in terms of a probability distribution, a second-order distribution, or a credal set. 
What we are interested in is a quantification of the epistemic and aleatoric (and maybe total) uncertainty associated with such representations. More formally, we are seeking a measure of epistemic uncertainty, $\eu$, and a measure of aleatoric uncertainty, $\au$. In the following, we discuss this problem for probabilistic and Bayesian agents, which prevail in the machine learning literature so far.

\subsection{Probabilistic Agents}

Recall that a probabilistic agent represents predictive uncertainty in terms of a distribution $\vtheta$ on $\cY$. 
The most well-known measure of uncertainty of a single probability distribution is the (Shannon) entropy, which, in our case of a discrete $\cY$, is given as 
\begin{equation}\label{eq:shannon}
S ( \vtheta ) \defeq   - \sum_{k=1}^K \theta_k \cdot \log_2 \theta_k  \, ,
\end{equation}
where $0 \log 0 = 0$ by definition. 
This measure is the most obvious candidate to quantify the aleatoric uncertainty of a probabilistic agent, i.e., $\au = S(\vtheta)$. As such an agent pretends to have precise knowledge about the predictive distribution, the epistemic uncertainty is 0. 

The Shannon entropy can be justified axiomatically, and different axiomatic systems have been proposed in the literature \citep{csis_ac08}.

\subsection{Bayesian Agents}
\label{sec:ba2}

The Bayesian perspective, according to which the epistemic state of the learner is represented by the posterior on the hypothesis space, is quite common in the machine learning community. Most recently, the problem of predictive uncertainty estimation has attracted specific attention in the field of deep neural networks. Corresponding methods typically seek to quantify (total) uncertainty on the basis of the predictive posterior distribution on $\cY$. Moreover, epistemic uncertainty is considered as a property of the posterior $q( \cdot \given \cD)$ resp.\ the induced distribution $p$ on $\Delta_K$: The less concentrated this distribution is, the higher the (epistemic) uncertainty of the learner. 

A principled approach to measuring and separating aleatoric and epistemic uncertainty, which is now widely applied in machine learning \citep{houlsby2011bayesian,depeweg2018decomposition,mobi_dc19}, is based on a classical information-theoretic result, according to which entropy (of a random variable $U$) can be decomposed into conditional entropy (of $U$ given another random variable $V$) and mutual information (of $U$ and $V$). In our case, considering the distribution on $\Delta_K$ as a random variable $\Theta$ (with distribution $p$) and the outcome as a random variable $Y$, we obtain the following:
\begin{align*}
    \on{TU} & = S(Y) = S(\bar{\vtheta})= - \sum_{k=1}^K \bar{\theta}_k \cdot \log_2 (\bar{\theta}_k) \, , \\[0.2cm]
    \on{AU} & = S(Y \given \Theta) = - \int p(\vtheta) \cdot S(\vtheta) \, d \, \vtheta \, , \\[0.2cm]
    \on{EU} & = S(Y) -  S(Y \given \Theta) = I(Y, \Theta) \, 
\end{align*}
where $I$ is mutual information.

\subsection{Levi Agents}
The Bayesian approach is clearly meaningful and has produced promising results in practice. Yet, it has recently been criticised for various reasons. Some conceptual problems of second-order distributions for representing the learner's belief have already been mentioned in Section \ref{sub:levi}. Moreover, \cite{wimmer2023quantifying} provide a critical discussion of the quantification of uncertainty in terms of mutual information and conditional entropy  (Section \ref{sec:ba2}), and demonstrate that these measures may show counter-intuitive behavior. Much of the problems revealed have to do with with ``averaging'' of (first-order) uncertainty over the learner's belief, i.e., over the second-order distribution. 

We take these criticisms as motivation to elaborate on the Levi agent as an alternative to the Bayesian approach, which essentially means replacing the second-order distribution by a credal set for representing the learner's epistemic state. More specifically, we propose a novel class of highly flexible uncertainty measures for credal sets.

In the uncertainty literature, there is quite some work on defining uncertainty measures for credal sets and related representations. Here, aleatoric and epistemic uncertainty are also referred to as \emph{conflict} (randomness, discord) and \emph{non-specificity}, respectively \citep{yage_ea83}. 
The standard uncertainty measure in classical possibility theory (where uncertain information is simply represented in the form of subsets $A \subseteq \cY$ of possible alternatives) is the Hartley measure \citep{hart_to28}

\begin{equation}\label{eq:hartley}
H(A) = \log( |A|)  \, ,
\end{equation}

Just like the Shannon entropy, this measure can be justified axiomatically\footnote{For example, see Chapter IX, pages 540--616, in the book by  \cite{reny_pt}.}. 

Given the insight that conflict and non-specificity are two different, complementary sources of uncertainty, and (\ref{eq:shannon}) and (\ref{eq:hartley}) as well-established measures of these two types of uncertainty, a natural question in the context of credal sets is to ask for a generalized representation 
\begin{equation}\label{eq:aggregate}
\on{TU}(C) = \on{AU}(C) + \on{EU}(C) \, ,
\end{equation}
where 
$\on{AU}$ is a generalization of the Shannon entropy, and $\on{EU}$ a generalization of the Hartley measure.

As for the non-specificity part in (\ref{eq:aggregate}), the following generalization of the Hartley measure to the case of graded possibilities has been proposed by various authors \citep{abel_an00}:
\begin{equation}\label{eq:gh}
\on{GH}(C) \defeq  \sum_{A \subseteq \cY} \on{m}_C(A) \, \log(|A|) \, ,
\end{equation}
where $\on{m}_C: \,  2^{\cY} \longrightarrow [0,1]$ is the M\"obius inverse of the capacity function $\nu :\, 2^{\cY} \longrightarrow [0,1]$ defined by
\begin{equation*}\label{eq:cap}
\on{\nu}_C(A) \defeq \inf_{\vtheta \in C} q(A) 
\end{equation*}
for all $A \subseteq \cY$, that is,
\begin{equation*}
\on{m}_C(A) = \sum_{B \subseteq A} (-1)^{|A \setminus B|} \nu_C(B) \, .
\end{equation*}
The measure (\ref{eq:gh}) enjoys several desirable axiomatic properties, and its uniqueness was shown by \cite{klir_ot87}.

The generalization of the Shannon entropy as a measure of conflict turned out to be more difficult. The upper and lower Shannon entropy play an important role in this regard: 
\begin{equation*}\label{eq:gg}
S^*(C) \defeq \max_{\vtheta \in C} S(\vtheta) \, , \quad
S_*(C) \defeq \min_{\vtheta \in C} S(\vtheta)
\end{equation*}
Based on these measures, the following disaggregations of total uncertainty (\ref{eq:aggregate}) have been proposed \citep{abel_dt06}:
\begin{align}
S^*(C) & = \big(S^*(C) - \on{GH}(C) \big)  + \on{GH}(C)  \label{eq:unc1} \\
S^*(C) & = S_*(C)  + \big(S^*(C) - S_*(C) \big)   \label{eq:unc2}
\end{align}
In both cases, upper entropy serves as a measure of total uncertainty, which is again justified on an axiomatic basis. In the first case, the generalized Hartley measure is used for quantifying epistemic uncertainty, and aleatoric uncertainty is obtained as the difference between total and epistemic uncertainty. In the second case, epistemic uncertainty is specified in terms of the difference between upper and lower entropy.

Nevertheless, a fully satisfactory representation of aggregate uncertainty in the form (\ref{eq:aggregate}), with all three measures having nice theoretical properties, has not yet been found for the case of credal sets. Both $S^*$ and $\on{GH}$ appear to be well justified and enjoy strong axiomatic properties. To a slightly lesser extent, this is also true for $S_*$ (this measure violates the property of monotonicity). However, those measures in (\ref{eq:unc1}--\ref{eq:unc2}) that are derived in terms of difference violate most of the desirable properties.

\section{Proper Scoring Rule-based Measures}
\label{sec:psr}
In this section, we propose a novel construction of uncertainty measures for Bayesian and Levi agents based on (strictly) proper scoring rules. This idea has already been mentioned by \citet{sale2023second}, who proposed it as a generalization of the variance-based measures introduced in that paper, and has (independently) also been elaborated on in the very recent paper by \citet{kotelevskiiPredictive2024}, albeit only for the Bayesian case. Before explaining it in more detail, let us first introduce some notation and revisit definitions and characterizations related to proper scoring rules. 

In the domain of probabilistic forecasting and decision-making, proper scoring rules are instrumental for the rigorous assessment and comparison of predictive models. These scoring rules, originating from the seminal works of \cite{savage1971elicitation} and further developed by \cite{gnei_sp05}, provide a mechanism to assign numerical scores to probability forecasts, rewarding accuracy and honesty in predictions. Proper scoring rules, such as the Brier score and the log score, are uniquely characterized by their \textit{properness}—a property ensuring that the forecaster's expected score is optimized only when announcing probabilities that correspond to their true beliefs. This property is essential for eliciting truthful and well-calibrated probabilistic forecasts, a critical aspect in various fields. Employing these scoring rules leads to more reliable forecasting models, which in turn improve the processes of decision-making in scenarios that rely on probabilistic forecasting. 

Recall that $\mathcal{Y}$ denotes a finite label space in our supervised learning framework. Further, we identify the class of all probability measures on the label space with the $(K-1)$-simplex $\Delta_K$. In the following we will denote the extended real line by $\bar{\mathbb{R}} \defeq \mathbb{R} \cup \{-\infty, +\infty\}$. 
A function defined on $\cY$ taking values in $\bar{\mathbb{R}}$ is $\Delta_K$-quasi-integrable if 
it is measurable with respect to $\sigma(\cY)$, a suitable\footnote{In the following, we assume without loss of generality $\sigma(\mathcal{Y}) = 2^{\mathcal{Y}}$.} $\sigma$-algebra on $\mathcal{Y}$, and is quasi-integrable with respect to all $\vtheta\in\Delta_K$. We assume scoring rules to be negatively oriented, thus taking a classical ML perspective where we wish to minimize the corresponding loss.

\begin{definition}
    A scoring rule is a function $\ell : \Delta_K \times \cY \to \bar{\mathbb{R}}$ such that $\ell(\hat{\vtheta}, \cdot)$ is $\Delta_K$-quasi-integrable for all $\hat{\vtheta} \in \Delta_K$. We further write 
    \begin{equation}\label{eq:eps}
    L_\ell(\hat{\vtheta}, \vtheta^*) = \mathbb{E}_{Y\sim\vtheta^*}[\ell(\hat{\vtheta},Y)]
    \end{equation}
    to denote the expected loss. A scoring rule is called proper if 
    \begin{align}
        L_\ell(\vtheta^*, \vtheta^*) \leq L_\ell(\hat{\vtheta}, \vtheta^*) \quad  \text{for all} \; \hat{\vtheta}, \vtheta^* \in \Delta_K.
        \label{def:psr}
    \end{align}
    The scoring rule is strictly proper, if \eqref{def:psr} holds with equality if and only if $\hat{\vtheta} = \vtheta^*$.
\end{definition}

It is well known that scoring rules and their corresponding expected losses can be decomposed into a \textit{divergence} term and an \textit{entropy} term \cite{gnei_sp05, kullNovel2015}: 
\begin{align*}
D_\ell(\hat{\vtheta}, \vtheta^*) & = L_\ell(\hat{\vtheta}, \vtheta^*) - L_\ell(\vtheta^*, \vtheta^*)  \\[0.2cm]   
H_\ell(\vtheta^*) & = L_\ell(\vtheta^*, \vtheta^*) 
\end{align*}
The former represents the expected loss, or risk, of predicting $\hat{\vtheta}$ when the ground truth is $\vtheta^*$, while the latter captures the expected loss that materializes even when the ground truth $\vtheta^*$ is predicted. This highlights the inherent connection to the quantification of epistemic and aleatoric uncertainty: $H_\ell(\vtheta^*)$ is the irreducible part of the risk, and hence relates to aleatoric uncertainty, whereas $D_\ell(\hat{\vtheta}, \vtheta^*)$ is the ``excess risk'' that is purely due to the learner's imperfect knowledge and could be reduced by improving that knowledge.  

In the following, we will further define the relation between (strictly) proper scoring rules and uncertainty quantification in Bayesian agents and Levi agents, showing how the currently used measures are related and how we propose to improve them.
\newpage
\subsection{Bayesian Agents}
\label{subsec:bayesnovel}
Recall that, in the Bayesian case, the learner holds belief in the form of a probability distribution $p$ on $\Delta_K$, and 
epistemic uncertainty is defined in terms of mutual information. The latter can also be written as follows:
\begin{align}
    \on{EU}(p) & = \mathbb{E}_{\vtheta \sim p}[D_\ell(\bar{\vtheta}, \vtheta)]  \label{eq:eunft} \\[0.2cm]
    & = \mathbb{E}_{\vtheta \sim p}[L_\ell(\bar{\vtheta}, \vtheta) - L_\ell(\vtheta, \vtheta)] \nonumber \\[0.2cm]
    & = \underbrace{\mathbb{E}_{\vtheta \sim p}[L_\ell(\bar{\vtheta}, \vtheta) ]}_{\text{TU}(p)} -  
    \underbrace{\mathbb{E}_{\vtheta \sim p} [ L_\ell(\vtheta, \vtheta)]}_{\text{AU}(p)} \label{eq:tuau}
\end{align}
where $L_\ell$ is instantiated with $\ell$ as the log-loss. That is, EU is the \emph{gain}\,---\,in terms of loss reduction\,---\,the learner can expect when predicting, not on the basis of the uncertain knowledge $p$, but only after being revealed the true $\vtheta$. Intuitively, this is plausible: The more uncertain the learner is about the true $\vtheta$ (i.e., the more dispersed $p$), the more it can gain by getting to know this distribution. 

The connection to proper scoring rules is also quite obvious:
\begin{itemize}
    \item Total uncertainty in (\ref{eq:tuau}) is the expected (log-)loss of the learner when predicting optimally ($\bar{\vtheta}$) on the basis of its uncertain belief $p$. It corresponds to the expectation (with regard to $p$) of the expected loss (\ref{eq:eps}) with $\ell$ the log-loss. Broadly speaking, we average the score of the prediction $\bar{\vtheta}$ over the potential ground-truths $\vtheta \sim p$. 
    \item Aleatoric uncertainty is the expected loss that remains, even when the learner is perfectly informed about the ground-truth $\vtheta$ before predicting. Again, we average over the potential ground-truths $\vtheta \sim p$.
    \item Epistemic uncertainty is the difference between the two, i.e., the expected loss reduction due to information about $\vtheta$. 
\end{itemize}
This definition is clearly meaningful, but as mentioned before, may not have desirable behaviour when instantiated with the log-loss. Therefore, we modify and extend the Bayesian approach as follows: First, we allow for loss functions other than log-loss. Second, we no longer assume that the agent fully commits to the expectation $\bar{\vtheta}$ as a prediction. Instead, it predicts at random according to $p$. Hence, our measures of uncertainty become:
\begin{align}
    \on{EU}(p) &= \bE_{\hat{\vtheta} \sim p}[\bE_{\vtheta \sim p}[D_\ell(\hat{\vtheta}, \vtheta)]] \label{eq:euft} \\[0.2cm]
    \on{AU}(p) &= \mathbb{E}_{\vtheta \sim p}[H_\ell(\vtheta, \vtheta)] \nonumber\\[0.2cm]
    \on{TU}(p) &= \on{AU}(p) + \on{EU}(p) = \bE_{\hat{\vtheta} \sim p}[\bE_{\vtheta \sim p}[L_\ell(\hat{\vtheta}, \vtheta)]] \nonumber
\end{align}
Compared to (\ref{eq:eunft}), the measure (\ref{eq:euft}) can be seen as a more faithful representation of the agent's belief and its (epistemic) uncertainty. According to the latter, uncertainty is defined in terms of the expected loss when the agent predicts $\vtheta$ according to its true belief $Q$. As opposed to this, (\ref{eq:eunft}) defines uncertainty in terms of the expected loss when the agent chooses the \emph{optimal} (loss-minimizing) prediction $\hat{\vtheta}$ (which is given by $\bar{\vtheta}$). As a result, the (first-order) prediction behavior of the agent may deviate from its (second-order) belief, and in essence, this is exactly what the theoretical results by \citet{bengs2022pitfalls} have shown.

\begin{table}[t!]
\centering
\caption{Proper scoring rules and their decomposition into aleatoric and epistemic uncertainty for the Bayesian agent.}
\begin{tabularx}{\textwidth}{XXl}
\toprule
Loss & Aleatoric & Epistemic \\ \midrule
log & $\mathbb{E}_{\vtheta \sim p}[S(\vtheta)]$ & $\bE_{\hat{\vtheta} \sim p}[\bE^{}_{\vtheta \sim p}[D_{KL}(\vtheta||\hat{\vtheta})]]$ \\[0.1cm]
brier &  $\mathbb{E}_{\vtheta \sim p}[1 - \sumK\theta^2_k]$ & $\bE_{\hat{\vtheta} \sim p}[\bE^{}_{\vtheta \sim p}[\sumK(\hat{\theta}_k - \theta_k)^2]]$ \\[0.1cm]
spherical & $\bE_{\vtheta \sim p}[1 - ||\vtheta||_2]$ & $\bE_{\hat{\vtheta} \sim p}[\bE^{}_{\vtheta \sim p}[||\vtheta||_2 - \sumK \hat{\theta}_k\theta_k / ||\vtheta||_2]]$ \\[0.1cm]
zero-one & $\bE_{\vtheta \sim p}[1-\max\theta_k]$ & $\bE_{\hat{\vtheta} \sim p}[\bE^{}_{\vtheta \sim p}[\max \theta_k - \theta_{k=\argmax\hat{\theta}_k}]]$ \\[0.1cm] \bottomrule
\end{tabularx}
\label{tab:bayesian}
\end{table}

Table \ref{tab:bayesian} has an overview of common proper scoring rules and the proposed decomposition into aleatoric and epistemic uncertainty for Bayesian agents. Using log-loss as an instantiation results in expected conditional entropy as a measure of aleatoric uncertainty and expected pairwise KL-divergence for epistemic uncertainty, which coincides with a recently proposed measure of uncertainty by \citet{schweighoferIntroducing2023}.


\subsection{Levi Agents}
\label{subsec:novel}
For Levi agents, we have no way of computing an expectation, thus we ``max'' over a (credal) set $C$ instead of averaging over a distribution and define epistemic uncertainty in terms of the maximal gain:
\begin{equation*}
\on{EU}(C)  :=  \max_{\hat{\vtheta} ,\vtheta \in C}  \, D_\ell(\hat{\vtheta}, \vtheta)  \, , 
\end{equation*}
with the $\ell$-divergence 
\begin{equation*}
D_\ell(\hat{\vtheta}, \vtheta) := L_\ell(\hat{\vtheta}, \vtheta) - L_\ell(\vtheta, \vtheta) \, .
\end{equation*}
The latter is the expected regret (excess risk) when predicting $\hat{\vtheta}$ although the ground-truth is $\vtheta$. 
For the aleatoric uncertainty, we obtain lower and upper bounds as follows:
\begin{align*}
    \underline{\on{AU}}(C) & := \underset{\vtheta \in C} \newinf\  H_\ell (\vtheta) \, ,  \\[0.2cm]
    \overline{\on{AU}}(C)  & := \underset{\vtheta \in C} \sup\  H_\ell(\vtheta) \, , 
\end{align*}
with $H_\ell$ the $\ell$-entropy of $\vtheta$ given by
\begin{equation*}
H_\ell (\vtheta) := \mathbb{E}_{Y \sim \vtheta} \, \ell(\vtheta, Y) \, .
\end{equation*}

Further, if we accept the idea of an additive aggregation of aleatoric and epistemic uncertainty into a measure of total uncertainty, we obtain the following lower and upper bound for the latter:
\begin{align*}
    \underline{\on{TU}}(C) & = \underline{\on{AU}}(C) + \on{EU}(C)  \, , \\[0.2cm]
    \overline{\on{TU}}(C)  & = \overline{\on{AU}}(C) + \on{EU}(C) \, . 
\end{align*}

Let us finally consider some exemplary instantiations of our family of measures, i.e., concrete measures that are obtained by fixing a loss function $\ell(\cdot , \cdot)$. First, for the special case of the log-loss, we recover lower and upper entropy for AU. Moreover, as $D_\ell$ is the KL-divergence, EU is the maximal KL-divergence between any pair of distributions $\hat{\vtheta}, \vtheta \in C$.

Allowing for other losses increases flexibility and allows one to capture uncertainty of different kind. For example, log-loss essentially captures uncertainty regarding the outcome $y$ eventually observed. From the perspective of the learner, however, this uncertainty might not be the most relevant one. Instead, the learner might be more interested in the uncertainty about the best \emph{decision} to make, i.e., the best prediction. These uncertainties are clearly not the same. For example, consider the case of binary classification and suppose that $C = \{ \vtheta = (\theta_{pos},\theta_{neg}) \, \vert \, 1/2 < \theta_{pos} \leq 1 \}$. In this case, it is clear that the learner should predict positive. In other words, there is no (epistemic) uncertainty about the right decision, although the uncertainty about the outcome is still rather high. An exemplary instantiation accounting for decision uncertainty is the zero-one-loss:
\begin{equation*}
 \ell (\vtheta, y) =
  \begin{cases}
    0, & \text{if } y = \argmax_{k} \theta_k  \\
    1, & \text{otherwise.}
  \end{cases} 
\end{equation*}
Table \ref{tab:credal} gives an overview of common proper scoring rules and the proposed decomposition into aleatoric and epistemic uncertainty for Levi agents.


\begin{table}[t!]
\centering
\caption{Proper scoring rules and their decomposition into aleatoric and epistemic uncertainty for the Levi agent.} 
\begin{tabularx}{\textwidth}{XXl}
\toprule
Loss & Aleatoric (upper\textbackslash lower) & Epistemic \\ \midrule
log & $\underset{\vtheta \in C}{\sup}\backslash\underset{\vtheta \in C}{\newinf} \ S(\vtheta)$ & $\underset{\hat{\vtheta}, \vtheta \in C}{\max}\ D_{KL}(\vtheta||\hat{\vtheta})$ \\
brier &  $\underset{\vtheta \in C}{\sup}\backslash\underset{\vtheta \in C}{\newinf}\ 1-\sumK \theta^2_k$ & $\underset{\hat{\vtheta}, \vtheta \in C}{\max}\ \sumK(\hat{\theta}_k - \theta_k)^2$ \\
spherical & $\underset{\vtheta \in C}{\sup}\backslash\underset{\vtheta \in C}{\newinf}\ 1 - ||\vtheta||_2$ & $\underset{\hat{\vtheta}, \vtheta \in C}{\max}\ ||\vtheta||_2 - \sumK \hat{\theta}_k\theta_k / ||\vtheta||_2$ \\
zero-one & $\underset{\vtheta \in C}{\sup}\backslash\underset{\vtheta \in C}{\newinf}\ 1-\max\theta_k$ & $\underset{\hat{\vtheta}, \vtheta \in C}{\max}\ \max \theta_k - \theta_{k=\argmax\hat{\theta}_k}$ \\ \bottomrule
\end{tabularx}
\label{tab:credal}
\end{table}

\section{Concluding Remarks}
\label{sec:conc}
In the light of recent literature on uncertainty quantification criticising uncertainty measures defined in the Bayesian settings and the lack of a unified approach to quantify uncertainty for second-order distributions (Bayesian agents) and credal sets (Levi agents) as representations, we propose novel uncertainty measures for the quantification of aleatoric and epistemic uncertainty based on proper scoring rules. 

We showed how the commonly used uncertainty measures in the Bayesian setting can be seen as an instantiation of proper scoring rule decomposition and extended this idea to overcome the aforementioned criticisms. Additionally, we extend this idea to the setting of Levi agents, which still lacks commonly accepted measures for the quantification of uncertainty. 

In future work, we will extend the formal analysis of the proposed measures and elaborate on their properties. We will also conduct empirical studies to see how the measures behave in different practical scenarios. 

\subsubsection*{Acknowledgements}
Yusuf Sale is supported by the DAAD program Konrad Zuse Schools of Excellence in Artificial Intelligence, sponsored by the Federal Ministry of Education and Research.

\bibliographystyle{apalike}
\bibliography{references}  

\end{document}